\documentclass{article}

\usepackage{ml4cps}

\usepackage[utf8]{inputenc} 
\usepackage[T1]{fontenc}    
\usepackage{hyperref}       
\usepackage{url}            
\usepackage{booktabs}       
\usepackage{amsfonts}       
\usepackage{nicefrac}       
\usepackage{microtype}      
\usepackage{cleveref}       
\usepackage{lipsum}         
\usepackage{graphicx}
\usepackage{doi}
\usepackage{booktabs}

\usepackage{color}

\title{Reconstructing Fine-Grained Network Data using Autoencoder Architectures with Domain Knowledge Penalties}
\date{\today}

\newif\ifuniqueAffiliation
\uniqueAffiliationtrue

\ifuniqueAffiliation 
\author{ Mark Cheung\\
	Peraton Labs\\
	\texttt{mark.cheung@peratonlabs.com} \\
	\And
	Sridhar Venkatesan \\
	Peraton Labs\\
	\texttt{svenkatesan@peratonlabs.com} \\
}
\else
\usepackage{authblk}

\newbox{\orcid}\sbox{\orcid}{\includegraphics[scale=0.06]{orcid.pdf}} 
\author[1]{%
	\href{https://orcid.org/0000-0000-0000-0000}{\usebox{\orcid}\hspace{1mm}First author}%
}
\author[1,2]{%
	\href{https://orcid.org/0000-0000-0000-0000}{\usebox{\orcid}\hspace{1mm}Second author}%
}
\affil[1]{Affiliation, Address}
\affil[2]{Affiliation, Address}
\fi

\renewcommand{\shorttitle}{Reconstructing Network Data with Autoencoder and Domain Knowledge}

\hypersetup{
pdftitle={Reconstructing Network Data with Autoencoder and Domain Knowledge},
pdfsubject={Reconstruction ML},
pdfauthor={Mark Cheung, Sridhar Venkatesan},
pdfkeywords={ForML, network packet reconstruction, domain-knowledge penalties, PCAP, autoencoder},
}

\begin{document}

\maketitle

\begin{abstract}
The ability to reconstruct fine-grained network session data, including individual packets, from coarse-grained feature vectors is crucial for improving network security models. However, the large-scale collection and storage of raw network traffic pose significant challenges, particularly for capturing rare cyberattack samples. These challenges hinder the ability to retain comprehensive datasets for model training and future threat detection. To address this, we propose a machine learning approach guided by formal methods to encode and reconstruct network data. Our method employs autoencoder models with domain-informed penalties to impute PCAP session headers from structured feature representations. Experimental results demonstrate that incorporating domain knowledge through constraint-based loss terms significantly improves reconstruction accuracy, particularly for categorical features with session-level encodings. By enabling efficient reconstruction of detailed network sessions, our approach facilitates data-efficient model training while preserving privacy and storage efficiency.

\end{abstract}


\section{Introduction}
Machine learning models have revolutionized network security by enabling fast and scalable inference from vast datasets of network traffic, host events, and intrusion alerts. However, these models often require large amounts of high-fidelity training data, which is difficult to maintain due to storage constraints, privacy concerns, and the sheer volume of network traffic. The problem is further exacerbated in cybersecurity, where attack instances are relatively rare, making it crucial to retain all available attack samples for future model improvements. Unfortunately, raw packet captures (PCAPs) are expensive to store and manage, leading to potential data loss that hinders retraining and adaptation of security models.

High-fidelity session-level datasets are essential for network telemetry, threat detection, and protocol analysis, enabling deep insights into attack patterns and network behavior. However, due to the impracticality of storing large-scale packet-level data, there is a pressing need for methods that can encode and reconstruct missing packet-level details from more compact, coarse-grained representations such as flow summaries.

To address this challenge, our project aims to develop machine learning models that can encode and reconstruct fine-grained network session data from aggregated features. Our approach leverages autoencoders, particularly transformer-based and recurrent architectures, to learn statistical patterns while incorporating domain-informed constraints to ensure protocol consistency and structured restoration. We hypothesize that domain-informed reconstruction on sessions of packets can restore packet-level details with high fidelity, facilitating richer network analysis while mitigating storage and privacy constraints.

\begin{figure}[!ht]
    \centering
    \includegraphics[width=\linewidth]{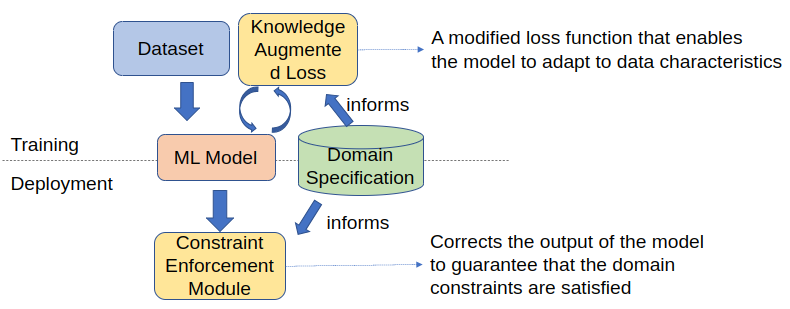}
    \caption{Overview of the proposed approach to encode and reconstruct fine-grained network sessions. Training is augmented with loss functions based on domain constraints and at deployment, the outputs of the model are modified to comply with the domain specification.}
    \label{fig:workflow}
\end{figure}

Fig.~\ref{fig:workflow} provides an overview of the proposed ML training and deployment approach to encode and reconstruct the fine-grained network sessions. This diagram depicts a machine learning system where domain knowledge is incorporated through an augmented loss function during training and constraints are enforced post-training to ensure model outputs align with specifications.

We structure the paper as follows. Section~\ref{sec:relatedwork} covers the background and related work. Section~\ref{sec:proposed_approach} details our approach to constructing the model with domain knowledge penalties. Section~\ref{sec:experimental_setup} explains the experimental setup. Sections~\ref{sec:results} presents our results. Finally, section~\ref{sec:conclusions} provides our conclusions and next steps.

\section{Related Work}
\label{sec:relatedwork}
Recent research in the field of data imputation has focused on developing models that reconstruct missing or aggregated data while adhering to inherent constraints. The Zoom2Net framework \cite{zoom2net} proposed a transformer-based architecture integrated with knowledge-augmented loss functions for time-series telemetry imputation, demonstrating improved consistency across categorical fields.

Moreover, various works have explored the use of deep learning architectures such as autoencoders for structured data reconstruction. Recurrent autoencoders and transformer-based models have proven effective in modeling sequential dependencies, with applications spanning traffic analysis and sensor data reconstruction. While transformer architectures offer long-range attention and scalability \cite{liu2025}, recurrent models provide robust modeling for short sequences with temporal coherence \cite{wang2018}.

Furthermore, several studies \cite{zhang} emphasize the importance of integrating domain knowledge into machine learning pipelines. By embedding formal rules or constraint-based penalties into loss functions, models achieve higher adherence to structural properties. These methods have been particularly effective in areas where protocol or schema compliance is non-negotiable, such as financial modeling and network protocol simulation.

In the context of network traffic analysis, deep learning has been increasingly utilized for tasks such as intrusion detection and flow prediction. Convolutional Neural Networks (CNNs) have been applied to analyze packet payloads and identify malicious patterns \cite{vinayakumar2019deep}. Graph Neural Networks (GNNs) have also shown promise in capturing the relational structure of network traffic, enabling effective anomaly detection and traffic classification \cite{wang2019graph}.

Specifically, for TCP flow reconstruction, prior work \cite{garfinkel} has explored statistical methods and rule-based systems to infer missing packet information. However, these methods often struggle with the complexity and variability of real-world network traffic. Deep learning models, particularly those leveraging sequence modeling capabilities, offer a more robust approach to handling such complexities.

Our approach combines these developments into a practical system for reconstructing PCAP session data, specifically focusing on TCP flows, using structured feature representations and domain-augmented loss functions to preserve protocol fidelity.

\section{Proposed Approach}
\label{sec:proposed_approach}

Our approach involves several key steps: data acquisition and preprocessing, autoencoder modeling, and loss function construction.

\subsection{Data Acquisition and Preprocessing}

We utilized packet capture (PCAP) files from the CyberVAN Cyberforce scenario \cite{chadha2016}. The preprocessing steps included:

\begin{enumerate}
    \item Parse packets using the Kamene library.
    \item Extract Ethernet, IP, and TCP headers (we focus on TCP only).
    \item Combine packets into sessions and store them for analysis
    \item Convert IP + Ethernet info into IP direction (since they are fixed for all packets in a session, except the direction of flow)
    \item Replace timestamp by time\_since first packet in each session
\end{enumerate}

We convert the byte sequences of the packet into numerical representations and in the default case, apply MinMax Scaler to all features, i.e., normalize all features between 0 and 1 by dividing them by the maximum minus minimum values. Using the domain knowledge, we later convert some numerical features into categorical encoding: 0 or 1 for binary, one-hot for few categories, and embedding vectors for many categories.
\subsection{Autoencoder Architecture}
\label{sec:autoencoder_arc}
\begin{figure}[!ht]
    \centering
    \includegraphics[width=\linewidth]{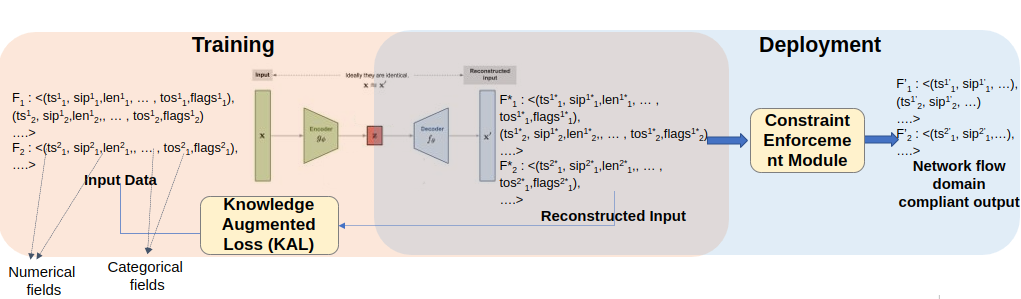}
    \caption{End-to-end architecture of the proposed framework}
    \label{fig:architecture}
\end{figure}

Fig.~\ref{fig:architecture} illustrates the end-to-end workflow of our proposed framework for reconstructing network telemetry from coarse-grained feature inputs using knowledge-augmented autoencoders and domain constraint enforcement.

During training, the model processes feature vectors $F_i$, which contain timestamped network flow records that include both numerical fields (e.g., packet lengths, timestamps) and categorical fields (e.g., source IP, type-of-service bits, TCP flags).

These feature vectors are processed by an encoder-decoder architecture. The autoencoder is trained with a Knowledge-Augmented Loss (KAL), which penalizes violations of known network semantics and relationships (such as monotonic timestamp progression or valid flag combinations), ensuring that the latent space captures both statistical patterns and domain-knowledge constraints.

The reconstructed input represents the output of the decoder, ideally closely matching the original feature representation while preserving structural flow relationships.

In the deployment stage, reconstructed flows are passed through a Constraint Enforcement Module, which ensures that the generated flows are not only statistically plausible but also domain-compliant with respect to known networking rules and flow structure. The output is a sequence of reconstructed telemetry that conforms to expected temporal ordering, header consistency, and valid protocol specifications.

\subsection{Loss Function with Domain Knowledge Integration}

We start with mean squared error (MSE) losses for all features. To ensure the reconstructed data preserves critical network properties, we designed a domain knowledge penalty function integrated into the loss function. This function includes:

\begin{itemize}
\item For features that are inherently binary, or those where only two distinct values are observed in our data samples, we treat them as binary variables. To train our model on these, we use the binary cross-entropy loss function.
\item Features with a limited number of categories, specifically those with fewer than ten, are encoded using one-hot vectors. We then employ the categorical cross-entropy (CE) loss to optimize the model's predictions for these features.
\item When dealing with some discrete features that have a large number of possible classes, such as port numbers, we represent them as embeddings. This allows the model to learn meaningful relationships between the classes, and we use categorical cross-entropy loss for training. The maximum number of ports is 65,535, but we only see 1041 different ports in the dataset (so we map them to 0 to 1040 and use an embedding dimension of 32). 
\end{itemize}

\section{Experimental Setup}
\label{sec:experimental_setup}
Here, we describe our experimental setup: model setup and training \& evaluation metrics.

\subsection{Model Setup}
As described in Section~\ref{sec:autoencoder_arc}, the model is designed to reconstruct network packets at the flow level, serving as a proof of concept for our imputation technique. We explore various types of autoencoders, including recurrent (LSTM, GRU, BiLSTM), multi-headed attention-based, and transformer autoencoders.

The input sequence length corresponds to the number of features—fewer when using numerical features only and more when incorporating one-hot encoding.

\subsection{Evaluation Metrics}
In our evaluation, we use both MSE and a binary reconstruction score to indicate whether  each reconstructed feature exactly matches its original value. The final results are derived by averaging the loss and score across packets (and features, when combined). We also present the scaled MSE loss, where we scale the result back to the original scale before calculating the MSE.

\subsection{Model Training}
We randomly split the data 80-20 for training and validation across sessions and report results based on the best validation loss. We use the ADAM optimizer with an initial learning rate of 0.001, a weight decay of 1e-5, and an early stopping patience of 100 epochs. The batch size is adjusted based on the model size (more for recurrent and less for transformer autoencoders).

\section{Results}
\label{sec:results}
In this section, we present experimental results demonstrating the effectiveness of our approach using domain knowledge. 

\subsection{Model comparison}

\begin{figure}[!ht]
    \centering
    \includegraphics[width=\linewidth]{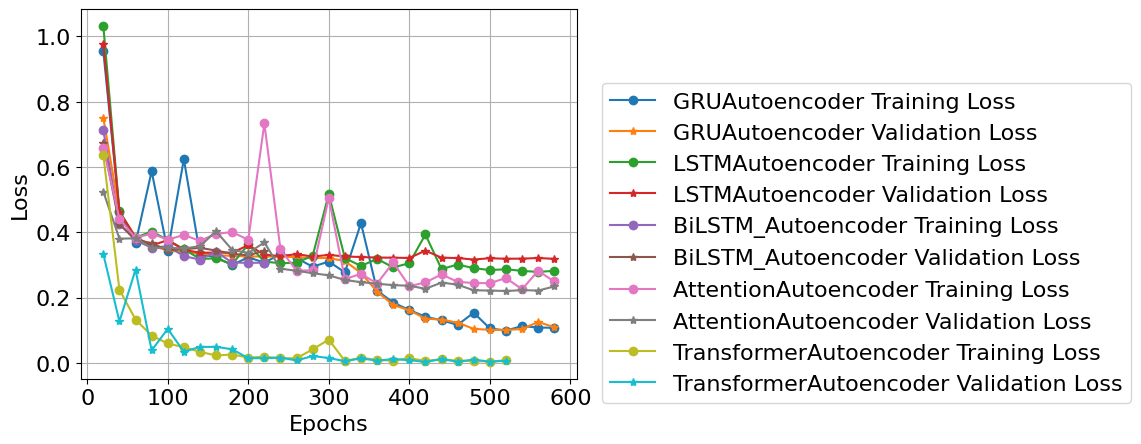}
    \caption{Model Comparison}
    \label{fig:model results comparison}
\end{figure}

Fig.~\ref{fig:model results comparison} shows the training and validation losses for GRU, LSTM, BiLSTM, attention, and transformer autoencoders. In general, the training and validation losses are similar, showing that no overfitting is occurring. However, there are some spikes during training, probably due in part to regularization. 

Among the architectures, transformer autoencoder has the lowest validation loss, stabilizing quickly. Attention autoencoder shows instability, with validation loss spikes. LSTM and GRU models converge more smoothly but at a higher loss.

\subsection{Domain Knowledge Loss}
\begin{figure}[!ht]
    \centering
    \includegraphics[width=\linewidth]{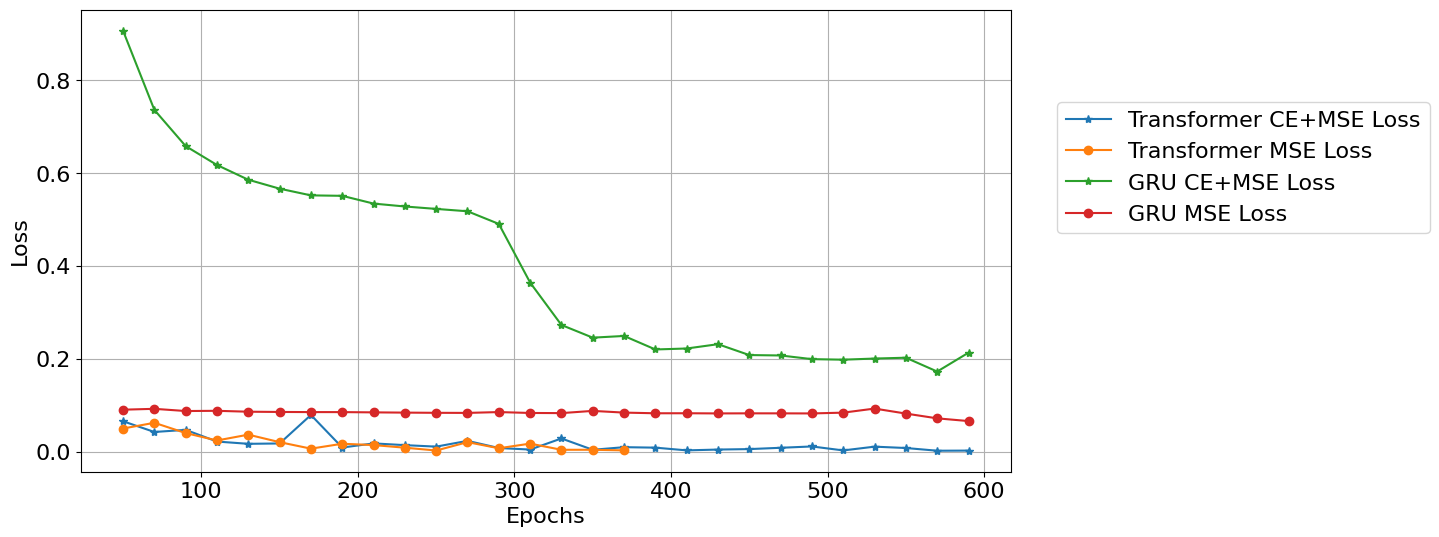}
    \caption{Comparison between loss functions. CE: Cross-Entropy loss , MSE: Mean-Squared Error loss}
    \label{fig:mse_vs_mse_ce}
\end{figure}
Fig.~\ref{fig:mse_vs_mse_ce} shows the comparison between loss functions with just MSE and MSE+CE. Transformer with CE+MSE loss outperforms others. GRU with CE+MSE loss shows a high initial loss but improves. MSE-only loss is more stable but converges slower. These plots do not clearly show that adding domain knowledge helps, so we look into the per-feature analysis.

\subsection{Per-Feature Analysis}

Table \ref{tab:mse_only} presents the performance metrics when only MSE loss is used. Most numerical features show high average reconstruction errors, with a clear correlation between the scaled MSE loss and the reconstruction error. In contrast, the binary features have low reconstruction errors, indicating they are easier to reconstruct accurately.

Table \ref{tab:combined_losses} shows the results when both Cross-Entropy (CE) and MSE losses are combined during training. Here, we notice a shift in the values for many features. While the numerical features still exhibit high average reconstruction errors and scaled MSE losses, the newly introduced one-hot features (with less than 10 categories) i.e., ip\_tos, tcp\_dataofs, and ip\_ttl as well as the embedded features i.e., source and destination ports now show zero or near-zero average reconstruction errors, indicating significant improvement in their reconstruction accuracy using domain knowledge. Note that since we use MSE for source and destination ports, their losses are initially much higher, so we trained the embedding separately first (simple feedfoward with embedding attains very low loss) before joining them with the rest of the model.

\begin{table}[!ht]
    \centering
    \setlength{\tabcolsep}{8pt}
    \caption{Per-Feature Results When Trained with MSE Loss Only}

    \begin{tabular}{lccc}
        \toprule
        \textbf{Numerical Feature} & \textbf{MSE Loss} & \textbf{Scaled MSE Loss} & \textbf{Avg. Reconstruction Error} \\
        \midrule
        tcp\_ack      & 0.0380  & 7.00E+17      & 1.0000  \\
        tcp\_seq      & 0.0313  & 5.76E+17      & 1.0000  \\
        src\_port     & 0.0195  & 80,504,472    & 1.0000  \\
        ip\_chksum    & 0.0181  & 77,395,760    & 1.0000  \\
        ip\_id        & 0.0141  & 60,411,184    & 1.0000  \\
        dst\_port     & 0.0639  & 264,136,768   & 0.9999  \\
        ip\_len       & 0.2935  & 625,534      & 0.9997  \\
        tcp\_window   & 0.0043  & 18,291,818    & 0.9995  \\
        payload\_size & 0.2917  & 616,641       & 0.9993  \\
        time\_since   & 0.0073  & 194,686       & 0.9984  \\
        ip\_ttl       & 0.0393  & 161.00        & 0.2562  \\
        ip\_tos       & 0.0175  & 4.4900        & 0.0429  \\
        tcp\_dataofs  & 0.0174  & 1.4100        & 0.0001  \\
        \midrule
        \textbf{Binary Feature} & \textbf{MSE Loss} & \textbf{Scaled MSE Loss} & \textbf{Avg. Reconstruction Error} \\
        \midrule
        tcp\_flag\_ECE   & 0.0287  & 0.0287  & 0.0000  \\
        tcp\_flag\_CWR   & 0.0211  & 0.0211  & 0.0000  \\
        ip\_direction    & 0.0187  & 0.0187  & 0.0000  \\
        tcp\_flag\_ACK   & 0.0172  & 0.0172  & 0.0000  \\
        tcp\_flag\_PSH   & 0.0112  & 0.0112  & 0.0000  \\
        ip\_ihl          & 0.0111  & 0.0111  & 0.0000  \\
        tcp\_flag\_SYN   & 0.0108  & 0.0108  & 0.0000  \\
        tcp\_flag\_RST   & 0.0100  & 0.0100  & 0.0000  \\
        tcp\_flag\_FIN   & 0.0054  & 0.0054  & 0.0000  \\
        ip\_proto        & 0.0046  & 0.0046  & 0.0000  \\
        ip\_frag         & 0.0046  & 0.0046  & 0.0000  \\
        ip\_flags        & 0.0045  & 0.0045  & 0.0000  \\
        tcp\_flag\_URG   & 0.0042  & 0.0042  & 0.0000  \\
        ip\_version      & 0.0032  & 0.0032  & 0.0000  \\
        tcp\_urgptr      & 0.0008  & 0.0008  & 0.0000  \\
        \bottomrule
    \end{tabular}
    \centering
    \label{tab:mse_only}
\end{table}

\begin{table}[!ht]
    \centering
    \renewcommand{\arraystretch}{1.2}
    \setlength{\tabcolsep}{10pt}
        \caption{Per-Feature Results When Trained with CE+MSE Losses}
    \begin{tabular}{lccc}
        \toprule
        \textbf{Numerical Feature} & \textbf{MSE Loss} & \textbf{Scaled MSE Loss} & \textbf{Avg. Reconstruction Error} \\
        \midrule
        tcp\_ack      & 1.1119  & 2.05E+19        & 1.0000 \\
        ip\_chksum    & 0.2506  & 1.07E+9         & 1.0000 \\
        tcp\_seq      & 0.1585  & 2.91E+18        & 1.0000 \\
        ip\_id        & 0.0815  & 350,156,544     & 0.9999 \\
        tcp\_window   & 0.1133  & 486,579,552     & 0.9999 \\
        time\_since   & 0.1395  & 3,744,877       & 0.9996 \\
        ip\_len       & 0.4842  & 1,032,158   & 0.9988 \\
        payload\_size & 0.5050  & 1,067,565     & 0.9981 \\
        \midrule
        \textbf{Binary Feature} & \textbf{Binary CE Loss} & \textbf{Scaled MSE Loss} & \textbf{Avg. Reconstruction Error} \\
        ip\_direction   & 0.0032  &	N/A             & 0.0000 \\
        tcp\_flag\_ECE & 0.0025  & N/A              & 0.0000 \\
        tcp\_flag\_ACK & 0.0015  & N/A              & 0.0000 \\
        tcp\_flag\_CWR & 0.0012  & N/A              & 0.0000 \\
        tcp\_flag\_URG & 0.0009  & N/A              & 0.0000 \\
        tcp\_flag\_PSH & 0.0009  & N/A              & 0.0000 \\
        tcp\_flag\_SYN & 9.15E-08 & N/A              & 0.0000 \\
        ip\_frag       & 9.15E-08 & N/A              & 0.0000 \\
        ip\_ihl        & 9.15E-08 & N/A              & 0.0000 \\
        ip\_flags      & 9.15E-08 & N/A              & 0.0000 \\
        ip\_version    & 0.0000  & N/A              & 0.0000 \\
        tcp\_flag\_RST & 0.0000  & N/A              & 0.0000 \\
        tcp\_urgptr    & 0.0000  & N/A              & 0.0000 \\
        ip\_proto      & 0.0000  & N/A              & 0.0000 \\
        tcp\_flag\_FIN & 0.0000  & N/A              & 0.0000 \\
        \midrule
        \textbf{Multi Categorical Feature} & \textbf{CE Loss} & \textbf{Scaled MSE Loss} & \textbf{Avg. Reconstruction Error} \\
        ip\_tos        & 0.0272  & N/A              & 0.000145 \\
        tcp\_dataofs   & 0.0030  & N/A              & 0.0000 \\
        ip\_ttl        & 0.0027  & N/A              & 0.0000 \\
        \bottomrule
        \textbf{Embedded Feature} & \textbf{MSE Loss} & \textbf{Scaled MSE Loss} & \textbf{Avg. Reconstruction Error} \\
        dst\_port     & 0.051  & N/A     & 0.0000 \\
        src\_port     & 0.049  & N/A     & 0.0000 \\
        \bottomrule
    \end{tabular}
    \centering
    \label{tab:combined_losses}
\end{table}

\subsection{Session- vs. Packet-level Encoding}
In our earlier experiments, we used session-level encoding by default. Here, we benchmark that against a simpler packet-level encoding using an autoencoder.

Our results show that packet-level reconstruction yields a slight improvement in accuracy for certain numerical features, such as ip\_len and payload\_size, where the average reconstruction errors drop to 0.7791 and 0.5823, respectively. However, for other features, the reconstruction error remains close to 0 or 1, indicating that the model predicts them perfectly or not at all. This reflects a broader trend: categorical fields are often reconstructed with high accuracy, while numerical fields tend to be more error-prone.

This leads us to the strength of session-level encoding, particularly in settings where data may be missing. Because sessions group multiple related packets, they provide richer context for inference. Missing fields in one packet can often be inferred from the presence of similar fields in other packets within the same session. In contrast, packet-level models must make predictions based on limited, isolated data—often defaulting to the most common value seen during training. This makes them less robust when faced with incomplete information.

\begin{figure}[!ht]
    \centering
    \includegraphics[width=\linewidth]{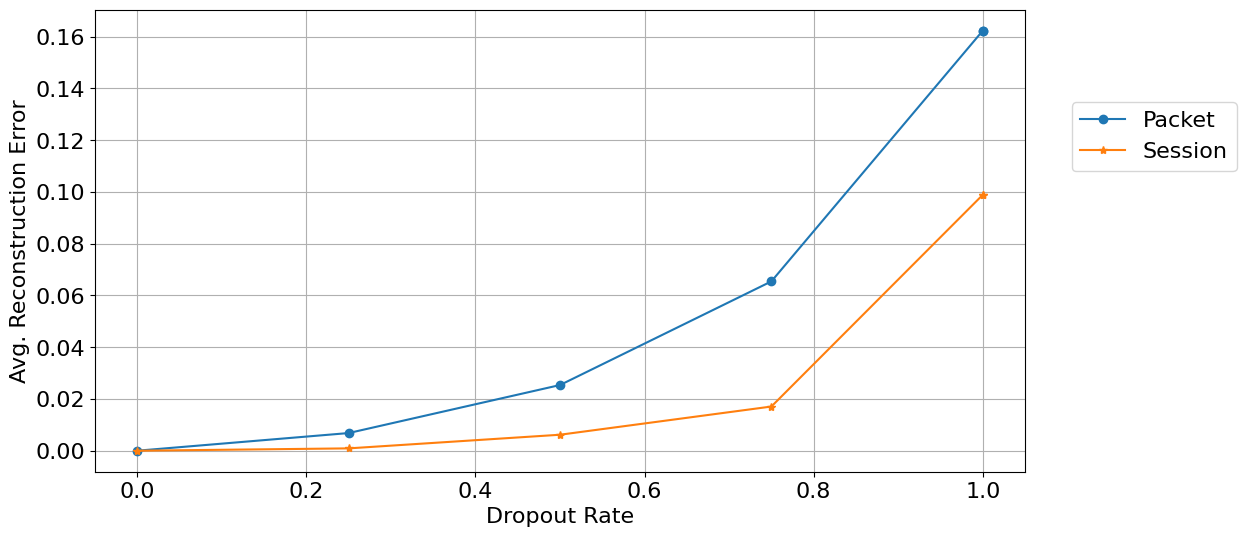}
    \caption{Reconstruction loss comparison between session-level and packet-level encoding with varying dropout rates.}
    \label{fig:session_vs_packet}
\end{figure}

To simulate missing fields, we trained both a packet-level autoencoder and a session-level transformer autoencoder, applying dropout after the input layer at rates of 0, 0.25, 0.5, 0.75, and 1. Dropout was used to mimic missing data in categorical features (numerical feature errors remained around 1 across all settings). Fig.~\ref{fig:session_vs_packet} shows that session-level encoding outperforms packet-level encoding in terms of reconstruction loss. While packet-level encoding performs slightly better with no dropout, its performance degrades more sharply as dropout increases. In contrast, session-level encoding remains more stable, with reconstruction loss rising more gradually—reaching only 0.0989 at the highest dropout rate—compared to 0.1623 for packet-level encoding. This difference stems from session-level encoding's ability to incorporate both packet count and contextual information, whereas packet-level encoding relies more heavily on isolated data points. Overall, these findings highlight the robustness of and motivation for session-level encoding in handling data.

\section{Conclusions and Future Work}
\label{sec:conclusions}
In conclusion, our results underscore the importance of incorporating domain-knowledge penalties into the autoencoder loss function, which plays a crucial role in preserving the structural integrity of reconstructed network packets. The application of feature engineering—such as careful scaling and encoding—substantially enhances model performance, ensuring that the network's inherent characteristics are better represented. Furthermore, session-based modeling proves to be a promising method for capturing complex dependencies within network traffic, offering potential for improved detection and reconstruction of network activities. These insights suggest that a combination of well-designed features and advanced loss function strategies can drive more accurate and reliable models for network traffic analysis.

Moving forward, we plan to refine our model by weighting features based on their contribution to loss and leveraging shared and independent properties within session-based packets. We will also explore values scaling with integer constraints as it will enable interpretability. We will also explore optimizing feature representation to represent high-cardinality fields. 

Our current transformer autoencoder setup is minimal, with limited hidden dimensions and layers. To improve performance, we will scale up model complexity by tuning hyperparameters and leverage additional GPU resources. Finally, to enhance applicability, we will extend our analysis to other types of packets such as UDP packets.

\section*{Acknowledgments}
This material is based upon work supported by Peraton Internal Research and Development funding. Any opinions, findings, conclusions or recommendations expressed in this material are those of the author(s) and do not necessarily reflect the views of Peraton Labs.


\begin{thebibliography}{1}

\bibitem{chadha2016}
R.~Chadha, T.~Bowen, C.~J. Chiang, Y.~M. Gottlieb, A.~Poylisher, A.~Sapello, C.~Serban, S.~Sugrim, G.~Walther, L.~M. Marvel, E.~A. Newcomb, and J.~Santos.
\newblock Cybervan: A cyber security virtual assured network testbed.
\newblock In {\em MILCOM 2016 - 2016 IEEE Military Communications Conference}, pages 1125--1130, 2016.

\bibitem{garfinkel}
S.~Garfinkel and M.~Shick.
\newblock {Passive TCP reconstruction and forensic analysis with tcpflow}.
\newblock Technical report, Naval Postgraduate School, September 2013.

\bibitem{zoom2net}
F.~Gong, D.~Raghunathan, A.~Gupta, and M.~Apostolaki.
\newblock Zoom2net: Constrained network telemetry imputation.
\newblock In {\em Proceedings of the ACM SIGCOMM 2024 Conference}, ACM SIGCOMM '24, page 764–777, New York, NY, USA, 2024. Association for Computing Machinery.

\bibitem{liu2025}
Z.~Liu, Y.~Xie, Y.~Luo, Y.~Wang, and X.~Ji.
\newblock Transeca-net: A transformer-based model for encrypted traffic classification.
\newblock {\em Applied Sciences}, 15(6), 2025.

\bibitem{vinayakumar2019deep}
R.~Vinayakumar, M.~Alazab, K.~P. Soman, P.~Poornachandran, A.~Al-Nemrat, and S.~Venkatraman.
\newblock Deep learning approach for intelligent intrusion detection system.
\newblock {\em IEEE access}, 7:41525--41550, 2019.

\bibitem{wang2019graph}
K.~Wang, Y.~Chen, A.~Khoshgozaran, and A.~K. Roy-Chowdhury.
\newblock Graph neural networks for anomaly detection in large-scale temporal graphs.
\newblock In {\em Proceedings of the 25th ACM SIGKDD international conference on knowledge discovery \& data mining}, pages 1775--1785, 2019.

\bibitem{wang2018}
T.~Wong and Z.~Luo.
\newblock Recurrent auto-encoder model for large-scale industrial sensor signal analysis.
\newblock In E.~Pimenidis and C.~Jayne, editors, {\em Engineering Applications of Neural Networks}, pages 203--216, Cham, 2018. Springer International Publishing.

\bibitem{zhang}
W.~Zhang, F.~Liu, C.~M. Nguyen, Z.~L. {Ou Yang}, S.~Ramasamy, and C.-S. Foo.
\newblock Training neural networks with classification rules for incorporating domain knowledge.
\newblock {\em Knowledge-Based Systems}, 294:111716, 2024.

\end{thebibliography}

\end{document}